\documentclass[12pt]{article}
\usepackage{chicago,alltt,epsf}

\def\pLeX{$\cal L$\kern-.3em\raise-.5ex\hbox{$\cal E$}\kern-.40em%
         \hbox{$\cal X$\kern-.24em\raise.21ex\hbox{{\it 4}}}}
\def\LeX{\protect\pLeX}

\def\trafo{{\sc Trafo}}

\begin{document}
\title{\bf A Lexical Semantic Database for Verbmobil\thanks{In: {\it
Proceedings of the 4\/$^{th}$ Conference on Computational Lexicography
and Text Research, COMPLEX 96, Budapest, 15-17 September 1996.}}
\thanks{The research
reported in this paper was supported by the German Bundesministerium
f\"ur Bildung, Wissenschaft, Forschung und Technologie under contract
01 IV 101 R and 01 IV 101 G6. We wish to thank our colleagues in the
Verbmobil project, especially Ronald Bieber, Johan Bos, Michael Dorna,
Markus Egg, Martin Emele, Bj\"orn Gamb\"ack, Gunter Gebhardi, Manfred
Gehrke, Julia Heine, Udo Kruschwitz, Daniela Kurz, Kai Lebeth,
Christian Lieske, Yoshiki Mori, Rita N\"ubel, Joachim Quantz, Sabine
Reinhard, Stefanie Schachtl, Michael Schiehlen, Feiyu Xu.}}

\author{ 
        {\sc Johannes Heinecke} \\ 
	Humboldt--Universit\"at zu Berlin \\
	Computerlinguistik \\
	J\"agerstra\ss{}e 10/11 \\
	D-10099 Berlin \\
	Germany \\
	{\small\tt heinecke@compling.hu-berlin.de} 
        \and
	{\sc Karsten L. Worm} \\
	Universit\"at des Saarlandes \\
	Computerlinguistik \\
	Postfach 15 11 50 \\
	D-66041 Saarbr\"ucken \\
	Germany \\
	{\small\tt worm@coli.uni-sb.de}
}
\date{}

\maketitle\thispagestyle{empty}

\begin{abstract}
This paper describes the development and use of a lexical semantic
database for the Verbmobil speech--to--speech machine translation
system. The motivation is to provide a common information source for the
distributed development of the semantics, transfer and semantic
evaluation modules and to store lexical semantic information
application--independently.

\bigskip
The database is organized around a set of abstract semantic classes
and has been used to define the semantic contributions of the lemmata
in the vocabulary of the system, to automatically create semantic
lexica and to check the correctness of the semantic representations
built up. The semantic classes are modelled using an inheritance
hierarchy. The database is implemented using the lexicon formalism
\LeX\ developed during the project.
\end{abstract}


\section{Introduction}

The distributed development of the modules of a large natural language
processing system at different sites makes interface definitions a
vital issue. It becomes even more urgent when several modules with
the same intended functionality are developed in parallel and should
be indistinguishable with respect to their input--output--behaviour.

Another important issue is the acquisition and maintenance of lexical
information which should be stored independently of an application to
make it (re)usable for different purposes.

This paper describes the design and use of the Verbmobil Semantic
Database which we developed in order to deal with these issues in
the area of lexical semantics in Verbmobil.

\section{The Verbmobil Project}

The Verbmobil project\footnote{Information about Verbmobil, such as
available reports, can be retrieved via the World Wide Web: {\tt
http://www.dfki.uni-sb.de/verbmobil/}.}
\shortcite{wahlster:93es,lud:96} aims at the development of a
speech--to--speech machine translation system for face--to--face
appointment scheduling dialogues.

The application scenario of Verbmobil is that a speaker of German and
a speaker of Japanese try to schedule an appointment. They communicate
mostly in English, which they understand better than they speak it. If
they they want to say something they cannot express in English, they can 
have the Verbmobil system translate from both their native languages to
English.

The system is being developed by about 30 partners from academia and
industry in Germany, the United States and Japan.  A first version,
the {\it Demonstrator}, was completed in early 1994; for autumn 1996
the release of the {\it Research Prototype\/} is scheduled, which
marks the end of the first project phase.
A second phase is expected
to start in 1997.

\begin{figure}
\epsfxsize=.8\textwidth

\centerline{\epsffile{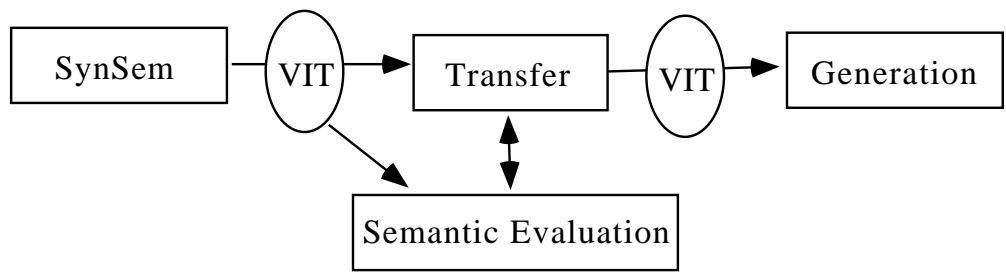}}

\caption{\label{archi}The relevant part of the Verbmobil architecture (simplified)}
\end{figure}

Verbmobil employs a semantic transfer approach to translation
\cite{de:96}, i.~e.  an input utterance is syntactically analyzed, a
semantic representation of the content is built up,\footnote{Syntactic
and semantic analysis proceed in parallel in the {\it Research
Prototype}, while they were two consequent processing steps in the
{\it Demonstrator}.} and this source language semantic representation
is mapped to a target language semantic representation by the transfer
module.  This representation is the input for the target language
generation. Additionally, a dialogue processing module and a semantic
evaluation module keep track of the discourse and answer disambiguation
queries. (The relevant part of the system architecture is shown in figure
\ref{archi}.)

\section{Motivation and Goals for the Semantic Database}

The architecture of Verbmobil makes it necessary for the semantics,
transfer, semantic evaluation and generation modules to agree on the
format and contents of the semantic representations they exchange.
E.~g.  the developers of the transfer module need to know how the
semantics of the different lemmata in the vocabulary is represented in
the structures produced by the syntax--semantics module ({\it
SynSem\/} for short), i.~e.  which predicates and structures they have
to map to the target language. On the other hand, semantics need to
know which readings have to be distinguished by transfer in order to
arrive at correct translations.

This need for information becomes even more urgent when, like in
Verbmobil, there are several SynSem modules (two for German, one
for Japanese), which have to produce compatible output, and the
different modules are developed independently and in parallel by
several partners at different sites.\footnote{In the following, we
concentrate on the Semantic Database for German. The Japanese version
follows the same principles.}

\begin{figure}
\begin{alltt}
vit( segment_description(ttestr4u1, yes,
\hfill     'wir machen einen termin aus'),
     [termin(l6,i2),                        % {\rm Semantics}
      ausmachen(l4,i1),
      decl(l5,h1),
      arg1(l4,i1,i3),
      arg3(l4,i1,i2),
      ein_card_qua(l3,i2,l1,h2,1),
      pron(l9,i3)],
     l5,                                    % {\rm Main Label}
     [s_sort(i1,ment_communicat_poly),      % {\rm Sorts}
      s_sort(i2,&(space_time,time_sit_poly)),
      s_sort(i3,&(human,person))],
     [prontype(i3,sp_he,std)],              % {\rm Discourse}
     [num(i3,pl),                           % {\rm Syntax}
      pers(i3,1),
      gend(i2,masc),
      num(i2,sg),
      pers(i2,3),
      cas(i2,acc),
      cas(i3,nom)],
     [ta_mood(i1,ind),                      % {\rm Tense and Aspect}
      ta_tense(i1,pres)],
     [ccom_plug(h2,l2),                     % {\rm Scope}
      ccom_plug(h1,l3),
      leq(l2,h2),
      leq(l2,h1),
      leq(l3,h1)],
     [pros_mood(l5,decl)],                  % {\rm Prosody}
     [sem_group(l2,[l4]),                   % {\rm Groupings}
      sem_group(l1,[l6])]
   )
\end{alltt}
\caption{\label{VIT-ex}A {\tt VIT} for {\it Wir machen einen Termin aus\/}
(``We arrange an appointment'').}
\end{figure}

As a frame for the exchange of semantic representations a common
format, the {\it Verbmobil Interface Term\/}, {\tt VIT} for short, has
been defined \cite{sc:96}. The {\tt VIT} is the central data structure
used at the interfaces between the language modules of Verbmobil. A
{\tt VIT} is a ten--place term with slots for an utterance identifier,
a list of labelled semantic predicates, a pointer to the most
prominent predicate, sortal, anaphoric and syntactic information,
temporal and aspectual properties, scope relations and prosodic
features. Figure \ref{VIT-ex} shows a {\tt VIT} for the sentence {\it
Wir machen einen Termin aus\/} (We arrange an appointment).

A {\tt VIT} is an underspecified representation for a set of discourse
representation structures \cite{kr:93} in which the scope of operators
is not fixed yet. In the example shown in figure \ref{VIT-ex} both the
scope of the declarative sentence mood operator, {\tt decl/2}, and of
the quantifier/indefinite, {\tt ein\_card\_qua/5}, are left
unspecified. They introduce {\it holes\/}, written as {\tt h1} and
{\tt h2}, as their scope, which can be {\it plugged\/} by structures
subordinated to them by means of less or equal constraints, written as
{\tt leq/2}. Different ways of plugging the holes result in different
readings. In addition to the {\tt leq/2} constraints determining all
possible readings, we supply a default scoping based on syntactic
structure in the predicates {\tt ccom\_plug/2}.\footnote{For more
details on this underspecified approach to semantics, the reader might
consult \shortcite{bos:95,lud:96}.}

All semantic predicates in the {\tt VIT} are labelled (their first argument
is the label). This allows us to group several predicates together
(using the {\tt sem\_group/2} predicate) and form complex
substructures which can occur in the scope of operators.

Apart from the purely semantic information mentioned so far, a {\tt
VIT} contains sortal constraints associated with discourse markers,
discourse information about anaphoric elements, syntactic agreement
and tense information. Since Verbmobil deals with spoken input, we
also represent prosodic information in the {\tt VIT}.\footnote{The
{\tt VIT} in figure \ref{VIT-ex} has been generated from typed input
and thus contains no real prosodic information.}

What is needed then in addition to the {\tt VIT} data structure
definition is a definition of the {\tt VIT}'s contents, for each lemma in
the vocabulary of the system a definition of the semantic predicates and
other types of information, e~g.  sortal restrictions, it introduces in
the slots of the {\tt VIT}. E.~g. for the verb {\it ausmachen\/} in the
example above, we need to specify that it introduces a predicate
\verb/ausmachen(L1,I1)/ together with argument roles
\verb/arg1(L1,I1,I2)/ and \verb/arg3(L1,I1,I3)/ in the semantics slot and
\verb/sort(I1,ment_communicat_poly)/ in the sorts slot.

If a source providing this kind of information to the developers of
the separate modules is available, the modules which deliver (the two
SynSem modules) or  process (especially the transfer module) {\tt
VIT}s conforming to this definition can be developed in parallel. It
would also be desirable to use this information source directly in the
construction of the linguistic knowledge bases to guarantee
consistency between the output and the specifications.

To meet these goals, we have developed the {\it Verbmobil Semantic
Database\/}, which we will describe in the remainder of this paper.


\section{Design and Implementation of the Database}

The semantic database is organized around a set of abstract semantic classes
\cite{sc:96}, which are used to classify the lemmata in the
vocabulary. It is implemented using the lexicon formalism \LeX.

\subsection{Semantic Classes} \label{ac}

The semantic classes in use are originally based on a
morpho--syntactic classification of the words in the vocabulary of the
system which has been refined to account for the semantic
properties. This has been decided upon, because words of a certain
word--class usually have the same semantic properties. In the example
given below, it is shown that transitive verbs all need an instance
and two arguments with their semantic/thematic roles.

For each semantic class a representation scheme, called the {\it
predscheme\/}, has been defined, which specifies the predicates together
with their arity and arguments appearing in a {\tt VIT} for
instances of the class.

As an example consider the class {\sf transitive\_verb\/}.  A transitive
verb is represented as {\tt R(L,I), argX(L,I,I1),
argY(L,I,I2)}.\footnote{{\tt X} and {\tt Y} stand for the values
$\{1,2,3\}$, since {\tt arg1, arg2, arg3} are the thematic roles used in
Verbmobil.}  I.~e., it introduces some relation \verb/R/ and two thematic
roles (\verb/I/ is the event variable, \verb/L/ a label used to refer to
the verb's semantic contribution, and \verb/I1/ and
\verb/I2/ are the instances filling the roles).  The verb's relation
and the thematic roles it assigns have to be defined for each verb in
the database. Cf. table \ref{class-table} for further examples of
semantic classes together with their predschemes.

\begin{table}
\begin{center}
\begin{tabular}{l|l|l}
Class			& PredScheme					& Example \\ \hline
\sf transitive\_verb	& {\tt R(L,I), argX(L,I,I1), argY(L,I,I2)}      & \it treffen \\
\sf common\_noun	& {\tt R(L,I)} 					& \it Termin \\
\sf det\_quant		& {\tt R(L,I,H)}				& \it jeder \\
\sf demonstrative	& {\tt demonstrative(L,I,L1)}			& \it dieser \\
\sf wh\_question	& {\tt whq(L,I,H), tloc(L2,I2,I1), time(L1,I1)}	& \it wann \\
\end{tabular}
\end{center}
\caption{\label{class-table}A few examples of semantic classes}
\end{table}


\subsection{The Lexicon Formalism \LeX}

The semantic database makes use of the lexicon formalism \LeX{}
developed in the course of the Verbmobil project \shortcite{vm-td19,gunter:96}.

The {\bf Lex}icon {\bf For}malism \LeX\ has been used since summer 1994
within Verbmobil's lexicon group.  It is based on feature-structures
(permitting disjunction and negation) embedded in an inheritance hierarchy
of classes. 

In \LeX\, the task of constructing a lexicon is split up into four parts:
\begin{enumerate}
\item Modelling the lexicon (i.e. its linguistic classes), 
\item data-acquisition (can be done at the same time by different contributors),
\item definition of the application-interface (data can be compiled into every format needed
after being processed by the
\LeX-machine), and 
\item efficient storage.
\end{enumerate}

Modelling a lexicon involves defining classes, their appropriate
features, and inheritance relations between classes. Examples for
defining classes will be given below in section \ref{scr};
appropriateness of features is dealt with in the remainder of this
section. For data acquisition, a graphical acquisition tool has been
implemented \cite{vm-td42}. How the application interface is used in the
context of the semantic database will be shown in section
\ref{appl}. Part of the application interface is the \LeX-\trafo\ which
outputs the stored information in any format required. A database system
for efficient storage has been developed \cite{udo:96}

Among other formalism constructs, the possible values of a feature can be
specified in two ways. If there is no restriction on the value of a
feature, it is assigned the {\it most general value} keyword ({\tt top}):

\begin{alltt}
   predname: \underline{top} .
\end{alltt}

Otherwise, the formalism allows to define the appropriateness conditions
of a feature, using disjunctions to specify the appropriate values as in
the following example (the underlined values are the appropriate ones which
can be assigned to the feature {\tt sort\_of\_inst}):

\begin{alltt}
   sort_of_inst: ( \underline{abstract} \verb+\+ \underline{anything} \verb+\+ \underline{communicat\_result\_poly} \verb+\+
                   \underline{communicat\_sit} \verb+\+ \underline{person} ) .
\end{alltt}

For constructing morphological lexica, inflection or lexical rules can
easily be implemented to generate multiple instances of a single
entry \shortcite{vm-m62,vm-m63}.

Database entries, called {\it bases\/}, are instances of a
class. Consequently, they assign values to the features they inherit
from their class which are not yet fully specified by the class
definition. For a verb's base, e.~g., one has to specify its predicate
name, thematic roles, the sort of its instance, etc.


\subsection{Semantic Classes and their Representation in \LeX} \label{scr}

The abstract semantic classes of section \ref{ac} have been modelled in
the lexicon formalism \LeX{} along the following lines.

Firstly, a general superclass {\tt semdb\_c} is defined from which all
classes inherit features for the lemma, the main predicate's name, the
part of speech etc.  The individual subclasses corresponding to the
abstract semantic classes additionally introduce a specific predscheme
for each predicate associated with words of this class and features for
sortal information, thematic roles etc.

\begin{small}
\begin{alltt}
   class semdb_c :< top >:    % {\rm - Main class from which}
                              % {\rm   all classes inherit.}
      syntax_link: top &      % {\rm - Link to syntactic lexicon.}
      predname: top &         % {\rm - Name of the semantic predicate.}
      lemma: top &            % {\rm - Lemma of the entry.}
      pos: top .              % {\rm - Part of Speech of the occurrences}
                              % {\rm   in the corpora.}
\end{alltt}
\end{small}

\begin{figure}
\epsfxsize=\textwidth
\epsffile{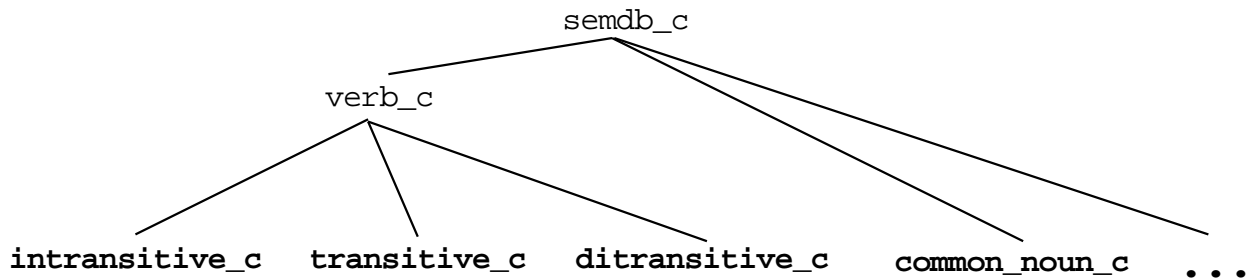}
\caption{\label{classes}Part of the class hierarchy}
\end{figure}

While the abstract semantic classes are not hierarchically organized,
their modelling in \LeX\ makes use of a hierarchy to capture
generalizations. 
For instance, we integrate all properties the verb classes
have in common and place them in an abstract verb class {\tt verb\_c}
from which all verb classes, e.~g. {\tt transitive\_c}, inherit,
cf. figure \ref{classes} (classes corresponding to semantic classes are
shown in boldface) and below.

\begin{small}
\begin{alltt}
   class verb_c :< semdb_c >:     % {\rm - All verbal classes inherit this.}
      sort_of_inst: top .         % {\rm - Sort of eventuality.}
 
     
   class transitive_c :< verb_c >:      % {\rm - Transitive verbs}
      semclass: transitive_verb &       % {\rm - Semantic class.}
      predscheme: 'L,I' &               % {\rm - PredScheme for the PredName}
                                        % {\rm   of all transitive verbs.}
      predscheme_a1: 'L,I,I1' &         % {\rm - PredScheme for the first}
      predscheme_a2: 'L,I,I2' &         % {\rm   and the second argument.}
      role_a1: (arg1 \verb+\+ arg2 \verb+\+ arg3) &   % {\rm - Thematic roles of the arguments}
      role_a2: (arg1 \verb+\+ arg2 \verb+\+ arg3) .   % {\rm   of the verb (restricted}
                                        % {\rm   to three valid values).}
\end{alltt}
\end{small}

As a second example, consider the following definition for the \LeX\
equivalent of the abstract semantic class {\sf common\_noun}:

\begin{small}
\begin{alltt}
   class common_noun_c :< semdb_c >:   % {\rm - Standard nouns}
      predscheme: 'L,I' &              % {\rm - PredScheme for standard nouns.}
      sort_of_inst: top &              % {\rm - Sort of instance.}
      semclass: common_noun .          % {\rm - Semantic class.}
\end{alltt}
\end{small}


\subsection{Representation of Lemmata}

A base for a lemma consists of its classification together with its
idiosyncratic properties in terms of feature values; it inherits the
feature values which are specified in the definition of the class.  Among
the idiosyncratic information we have predicate names, sortal
restrictions etc. Thus an entry inherits the predscheme from the class,
while the concrete predicate name in the predscheme is defined in the
entry itself.

\begin{small}
\begin{alltt}
   base 'Termin' :<< common_noun_c >>:     % {\rm - The entry `}Termin{\rm'}
                                           % {\rm   inherits its structure from}
                                           % {\rm   from the class `}common_noun_c{\rm'.}
      pos: 'NN' &                          % {\rm - Further individual}
      lemma: 'Termin' &                    % {\rm   specification for}
      syntax_link: 'termin' &              % {\rm   the current entry.}
      predname: 'termin' &
      sort_of_inst: 'time_sit_poly' .


   base 'ausmachen' :<< transitive_c >>:   % {\rm - The entry `}ausmachen{\rm'}
                                           % {\rm   inherits its structure from}
                                           % {\rm   the class `}transitive_c{\rm'.}
      pos: 'VVFIN;VVINF' &                 % {\rm - Further specifications.}
      lemma: 'ausmachen' &
      syntax_link: 'ausmachen' &
      predname: 'ausmachen' &
      sort_of_inst: (communicat_sit \verb|\| mental_sit) &
      role_a1: 'arg1' &
      role_a2: 'arg3' .
\end{alltt}
\end{small}

When processing the class definitions and the bases, the \LeX-machine
will calculate all instances from the specifications and expand the base
accordingly.


\section{Application of the Semantic Database} \label{appl}

The Semantic Database is currently being used for creating the semantic
lexica of the syntactic--semantic modules of Verbmobil, for producing a
table of lemmata with the predicates and other types of information they
introduce in a {\tt VIT} and for checking the correctness of the
generated interface terms automatically; it can also be accessed via the World
Wide Web.

A similar procedure is used to generate the semantic lexicon etc. for the
Japanese syntactic--semantic module of Verbmobil \cite{mori:96}.


\subsection{Creation of the Semantic Lexicon}

Consider the compilation of the semantic lexicon from the database for
the German SynSem module SynSemS3.\footnote{SynSemS3 is the
syntactic--semantic module developed by {\bf S}iemens AG (syntax),
University of the {\bf S}aarland and University of {\bf S}tuttgart
(semantics). The other SynSem module developed by IBM Germany makes
use of the table output (cf. section \ref{table}) of the database to
create a semantic lexicon.} To guarantee consistency between the
output of the SynSem module and the specifications in the database,
the semantic lexicon is generated out of the semantic database.

After the \LeX--machine has processed the entries and expanded them 
according to the class definitions, the \LeX--\trafo{} compiles the 
\LeX\ output into the format required for the semantic lexicon.

\begin{small}
\begin{alltt}
    sort1_trafo(Base, Class,          % {\rm - Default rule for entries}
      [ predname:Predn,               % {\rm   with one sort.}
        syntax_link:Sl,
        sort_of_inst:Si,
        usb_macro:M
      ] ) => 
    fmt(\verb+"+sem_lex(Cat, ~w) short_for~n     ~w(Cat, ~w, (~w)) .~n\verb+"+, 
       [Sl, M, Predn, Si], []).


    trans_trafo(Base, Class,          % {\rm - Rule for bivalent verbs.}
      [ predname:Pn,
        syntax_link:Sl,
        sort_of_inst:Si,
        role_a1:R1,
        role_a2:R2,
        usb_macro:M
      ] ) => 
    fmt("sem_lex(Cat, ~w) short_for~n    ~w(Cat, ~w, (~w), [~w,~w]) .~n", 
       [Sl, M, Pn, Si, R1,R2], []).
\end{alltt}
\end{small}

The two examples above appear in the semantic lexicon as:

\begin{small}
\begin{alltt}
sem_lex(Cat, termin) short_for
     common_noun_sem(Cat, termin, (time_sit_poly)) .
sem_lex(Cat, ausmachen) short_for
     trans_verb_sem(Cat, ausmachen, (communicat_sit;mental_sit), 
                    [arg1,arg3]) .
\end{alltt}
\end{small}

The syntactic lexicon contains calls to the macro {\tt sem\_lex/2} which
is expanded in the semantic lexicon as shown above. The mapping from
syntactic to semantic lexical entries is achieved via the second argument
of {\tt sem\_lex/2}, which originates from the feature {\tt syntax\_link}
in the semantic database.\footnote{The first argument of {\tt sem\_lex/2}
ranges over entry nodes of the feature structures of the lexical entry.}

\subsection{Table--based Representation} \label{table}

Apart from compiling out semantic lexica, we generate a table of lemmata
together with their semantic representations and additional information
out of the database by using a different set of transformation rules for
\LeX-\trafo{}. This table is used by the transfer developers as a
basis for writing transfer rules and as an information source for the
automatic correctness check on {\tt VIT} representations.

\begin{small}
\begin{alltt}
   transitive_trafo(Base, Class,      % {\rm - Rule for bivalent verbs.}
     [ lemma:Lm,
       pos:Pos,
       semclass:Semc,
       predname:Pn,
       predscheme:Ps, 
       predscheme_a1:Ps1, 
       predscheme_a2:Ps2,
       role_a1:Ra1,
       role_a2:Ra2,
       sort_of_inst:Si,
       inst_link:Il,
       sort_a1:Sa1, a1_link:Al1,
       sort_a2:Sa2, a2_link:Al2
     ] ) => 
   fmt(\verb+"+~w ~w ~w ~w,~w,~w ~w ~w(~w),~w(~w),~w(~w) ~w/~w,~w/~w,~w/~w - -~n\verb+"+, 
      [ Base, Lm, Pos, Pn,Ra1,Ra2, Semc, Pn,Ps, Ra1,Ps1, Ra2,Ps2, 
        Il,Si,Al1,Sa1,Al2,Sa2], []).


   default_ps1_inst1(Base, Class,         % {\rm - Default rule for entries with}
     [ lemma:Lm,                          % {\rm   one PredScheme and one Sort}
       pos:Pos,                           % {\rm   (used e.g. by `}common_noun{\rm').}
       semclass:Semc, 
       predname:Pn, 
       predscheme:Ps,
       sort_of_inst:Si
     ] ) =>
   fmt(\verb+"+~w ~w ~w ~w ~w ~w(~w) ~w  - -~n\verb+"+, 
     [ Base, Lm, Pos, Pn, Semc, Pn,Ps, Si], []).
\end{alltt}
\end{small}

In the table output the two examples above appear as:

\begin{small}
\begin{alltt}
Termin Termin NN termin common_noun termin(L,I) I/time_sit_poly - -
ausmachen ausmachen VVFIN;VVINF ausmachen,arg1,arg3 transitive_verb {\dots}
\hfill ausmachen(L,I),arg1(L,I,I1),arg3(L,I,I2) I1/communicat_sit;mental_sit - -
\end{alltt}
\end{small}

In general the concept of \trafo\ is trying to map the output of the
\LeX-machine onto the first matching rule in the rule system. Thus
only a few class specific rules are necessary, default rules
will cover the entries of the majority of the classes to be
transformed.


\section{Summary}

We have successfully used the semantic database to deal with about 2000
German and 150 Japanese lemmata for version 1.0 of the Research Prototype
in the way described, especially to generate semantic lexica for the
German syntax--semantics module SynSemS3, and the Japanese one developed
by DFKI Saarbr\"ucken and the University of the Saarland. 

The use of the semantic database by both the semantics module and the
transfer module guarantees consistency between the representations
produced by the semantics module and the expectations of the transfer
module, while both can be developed in parallel.


\bibliographystyle{chicago}
\bibliography{complex1}

\end{document}